\documentclass[10pt,twocolumn,letterpaper]{article}

\usepackage[pagenumbers]{iccv} % To force page numbers, e.g. for an arXiv version

\definecolor{iccvblue}{rgb}{0.21,0.49,0.74}
\usepackage[pagebackref,breaklinks,colorlinks,allcolors=iccvblue]{hyperref}
\usepackage{booktabs}
\usepackage{xcolor}

\def\paperID{*****} % *** Enter the Paper ID here
\def\confName{ICCV}
\def\confYear{2025}

\title{Glance-MCMT: A General MCMT Framework with Glance Initialization and Progressive Association}

\author{Hamidreza Hashempoor\\
{\tt\small hashemp.hamidreza@gmail.com}
}

\begin{document}
\maketitle
\begin{abstract}
We propose a multi-camera multi-target (MCMT) tracking framework that ensures consistent global identity assignment across views using trajectory and appearance cues. The pipeline starts with BoT-SORT-based single-camera tracking, followed by an initial glance phase to initialize global IDs via trajectory-feature matching. In later frames, new tracklets are matched to existing global identities through a prioritized global matching strategy. New global IDs are only introduced when no sufficiently similar trajectory or feature match is found. 3D positions are estimated using depth maps and calibration for spatial validation.
\end{abstract}    
\section{Introduction}
\label{sec:intro}

Multi-camera multi-target (MCMT) tracking aims to assign consistent object identities across disjoint camera views, where targets may appear in different cameras at different times. This problem becomes especially challenging in crowded scenes or when camera views have minimal overlap. In this work, we present a structured and modular MCMT pipeline that addresses the full tracking pipeline—from detection to global identity assignment—through a series of carefully designed stages.

Our approach begins by training a custom object detector to better handle a wider variety of object classes relevant to the target environment. This improves detection quality, particularly for underrepresented classes, and provides a more robust base for downstream tracking.

Next, we apply single-camera tracking using BoT-SORT, a well-known open-source tracking algorithm that combines motion prediction and appearance feature extraction. The output is a sequence of per-camera local tracklets. To further refine these results, we conduct post-processing of the single-camera tracklets in two ways:
\begin{itemize}
    \item We apply \textit{non-maximum suppression (NMS)} to merge overlapping tracklets that likely belong to the same object.
    \item We also perform \textit{representative feature } selection by clustering appearance features within each tracklet to extract stable embeddings, which are later used for global ID matching.
\end{itemize}

Once local tracking results are stabilized, we enter the global identity assignment stage, starting with a \textit{glance phase}. In this step, we analyze only the first few frames from all cameras to identify the globally unique set of objects. Using camera calibration matrices, we project 2D bounding boxes into real-world coordinates and derive the \textit{initial trajectory} for each tracklet. Tracklets are compared across views using: (i) Euclidean distance between real-world trajectories, and (ii) Feature similarity using a re-identification (ReID) model.

\begin{figure}[t]
    \centering
    \includegraphics[width=\columnwidth]{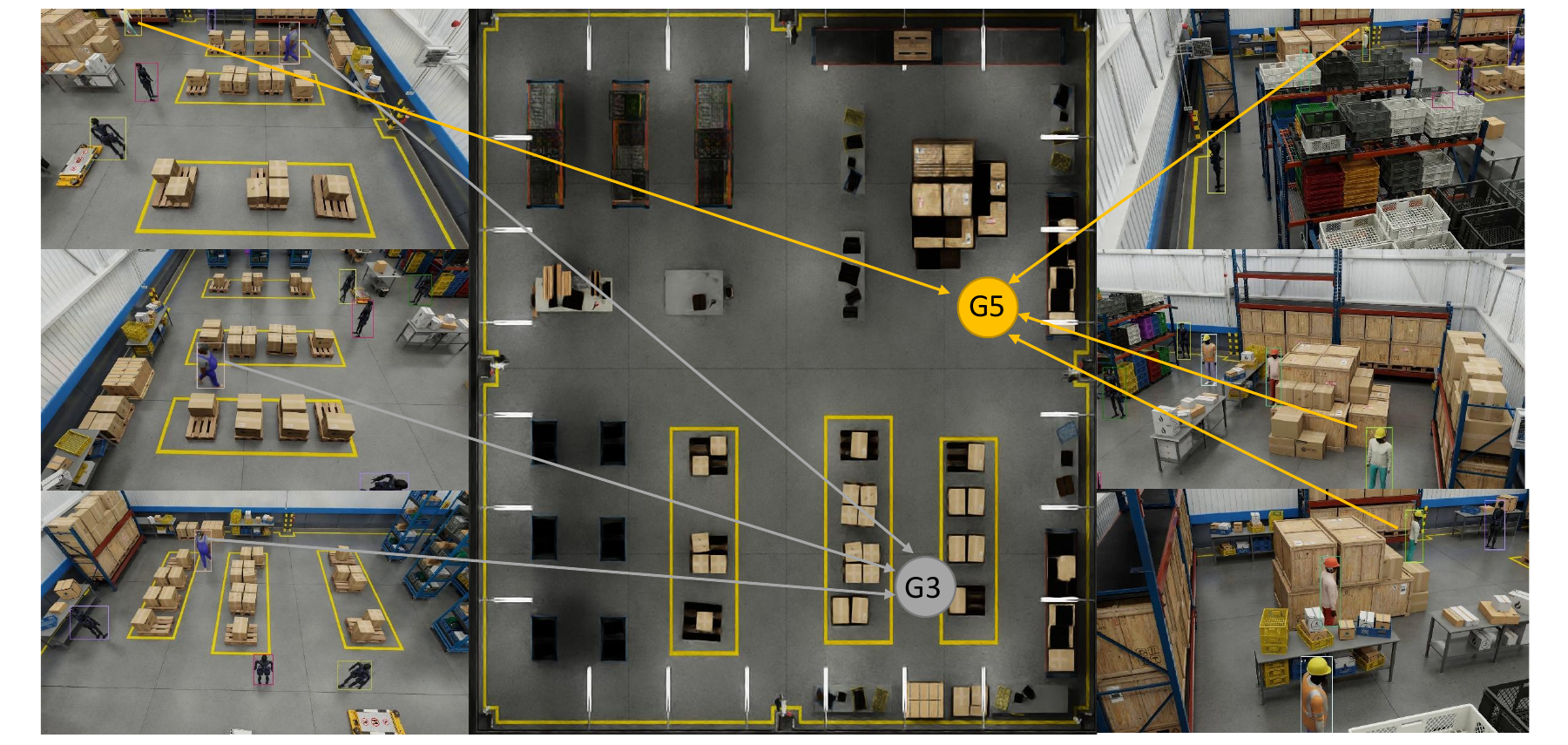}
    \caption{Overview of the proposed MCMT pipeline. Objects detected in each camera view are progressively associated and assigned consistent global IDs across views.}
    \label{fig:overview}
\end{figure}

Tracklets deemed similar based on both trajectory and feature cues are merged and assigned a global identity, which remains fixed for the remainder of the sequence.

For assigning the remaining unassigned tracklets, we use a \textit{progressive matching strategy}:
\begin{enumerate}
    \item At each step, we collect unassigned tracklets that \textit{overlap in time} with existing global tracklets.
    \item We compare these candidates with global identities based on trajectory and appearance similarity.
    \item If a strong match is found, the unassigned tracklet is added to the corresponding global ID.
    \item This process continues iteratively, \textit{growing the global tracklets} and shrinking the set of unassigned tracklets, until all tracklets are matched.
\end{enumerate}

This design allows us to flexibly introduce new global IDs only when existing global trajectories fail to match new observations under strict similarity conditions. The overview of the proposed MCMT is in Figure \ref{fig:overview}.

\vspace{0.3cm}
\noindent
Overall, our main contributions are:
\begin{itemize}
    \item \textbf{End-to-End Pipeline:} We propose a full multi-camera tracking system, starting from improved detection, single-camera tracking, to global identity assignment.
    \item \textbf{Tracklet Merging:} We refine single-camera outputs with NMS and spatial IoU-based merging to reduce fragmentation.
    \item \textbf{Trajectory-Based Global ID Assignment:} We leverage camera calibration to project bounding boxes to real-world space and match initial tracklets across views.
    \item \textbf{Progressive Global Matching:} Our iterative matching strategy efficiently associates remaining tracklets while preserving identity consistency.
    \item \textbf{Controlled Global ID Growth:} New global IDs are introduced only under strict conditions, minimizing identity inflation.
\end{itemize}
\section{Related Works}
\label{sec:related_works}

MCMT tracking has attracted significant attention across a range of applications, such as vehicle surveillance \cite{zhang2022dino} and pedestrian monitoring~\cite{huang2023enhancing,jeon2023leveraging,kim2023addressing,ghalamzan2021deep, hashempour2020data, li2023hierarchical,nguyen2023multi,specker2023reidtrack,yang2023integrating}. Despite differences in domain, typical MCMT pipelines follow a common structure: they begin with object detection and single-camera tracking, followed by inter-camera identity association to achieve consistent tracking across views.

\subsection{Object Detection}

 In MCMT tracking, the quality of object detection significantly impacts overall performance, as it directly influences the precision of both appearance and spatial localization features. A wide range of detection models have been explored to support real-time processing. These include one-stage architectures such as the You Only Look Once (YOLO) family~\cite{kim2023addressing,li2022yolov6,wang2023yolov7}
Among recent one-stage detectors, YOLOX~\cite{ge2021yolox} stands out for its decoupled head design, which separates the classification and localization branches to improve convergence and accuracy. It also adopts anchor-free detection and advanced training strategies for label assignment. Building on this foundation, several variants have extended YOLOX for multi-task scenarios. For example, some works introduce additional heads for extracting feature embeddings used in color, style, or direction classification~\cite{hashempoor2024featuresort}.

There are also two-stage detectors that rely on Region Proposal Networks (RPN), such as the R-CNN series~\cite{girshick2015fast,girshick2014rich,he2017mask,ren2015faster}. More recently, transformer-based detectors—such as Detection with Transformers (DETR) and its variants~\cite{carion2020end,zhang2022dino,zhu2020deformable}—have achieved state-of-the-art results, offering end-to-end detection without relying on hand-crafted proposals.

Joint Detection and Embedding (JDE) models~\cite{zhang2021fairmot,you2023utm} provide a one-shot alternative to traditional multi-stage pipelines by jointly predicting object locations and appearance embeddings, thereby improving runtime efficiency. However, this integration increases training complexity, as it demands both dense bounding box annotations and large-scale Re-ID data. Moreover, the effectiveness of JDE in re-identification remains uncertain, since evaluations are often limited to single-camera MOT datasets such as MOT17~\cite{dendorfer2021motchallenge} and MOT20~\cite{dendorfer2020mot20}. Finally, confidence threshold tuning is crucial: low recall can underestimate object counts, while low precision can lead to overestimation.

\subsection{Person Re-Identification}

Re-ID~\cite{ye2021deep} aims to recognize the same individual across different images using only visual cues. Similarity between identities is typically measured using learned feature embeddings. Early methods explored descriptor learning~\cite{gray2008viewpoint,schwartz2009learning} and color calibration techniques~\cite{javed2005appearance}, which remain foundational in many pipelines~\cite{bedagkar2014survey}. However, recent advances have been driven by representation learning-based models that extract discriminative embeddings~\cite{hadsell2006dimensionality,hoffer2015deep,deng2019arcface}.

These models are trained using loss functions that enforce high similarity between matching pairs and low similarity between mismatched pairs. Modern Re-ID architectures such as OSNet~\cite{zhou2019omni} and those trained on large-scale datasets like LUPerson~\cite{fu2021unsupervised} have demonstrated strong performance, achieving a mean Average Precision (mAP) above 0.85 on the widely used Market1501 benchmark~\cite{zheng2015scalable}.

\subsection{Single-Camera Multi-Target Tracking}

Single-camera multi-target tracking (SCMT) involves associating detected objects across consecutive frames to form coherent trajectories. While much of the literature has focused on pedestrian tracking, the general principles extend naturally to multi-class tracking scenarios involving other object categories. In SCMT, associations are typically made between bounding boxes representing the same physical object across time.

Regardless of the object category, data association methods can generally be categorized into motion-based and appearance-based approaches.

\subsubsection{Motion-Based Tracking}

Motion-based tracking relies on the assumption that object positions exhibit spatiotemporal continuity. Bounding box positions predicted from prior frames are matched to current detections using motion models such as the Kalman filter or in general Gaussian SSMs \cite{hashempoorikderi2024gated}. For example, SORT~\cite{bewley2016simple} uses a Kalman filter to estimate the object state and assigns detections based on Intersection over Union (IoU) with predicted boxes. Extensions like SimpleTrack~\cite{li2022simpletrack} employ generalized IoU to handle non-overlapping detections, while StrongSORT~\cite{du2023strongsort} improves temporal consistency by incorporating momentum-based predictions over longer frame windows.

Although motion cues are effective in single-camera settings, their utility in multi-camera tracking is limited. Motion-based association depends on spatial continuity, which breaks down across disjoint camera views. In such cases, further processing—such as coordinate transformation into a shared world space—is required to enable motion-based correspondence across cameras.

\subsubsection{Appearance-Based Tracking}

Appearance-based tracking operates on the assumption that objects maintain consistent visual characteristics over time. By extracting feature embeddings from detected instances, objects across frames can be associated if their visual similarity exceeds a predefined threshold. Unlike motion-based methods, appearance-based tracking does not rely on positional continuity, making it useful for scenarios with abrupt motion, occlusion, or missed detections.

Appearance-based strategies are applicable in both single-camera and multi-camera multi-object tracking. However, the similarity threshold for associating objects may need to be adapted to the context. In single-camera settings, visual continuity across adjacent frames helps maintain high appearance similarity. In contrast, multi-camera tracking often involves view changes, lighting differences, or partial occlusions, which can lower the visual similarity even for the same object, thus requiring more robust feature extraction and matching strategies.

\subsection{Multi-Camera Multi-Target Tracking}

MCMT tracking expands the traditional focus on pedestrian tracking to include a broader set of object categories. The 2025 AI City Challenge Track 1 emphasizes this more general tracking task across multiple overlapping camera views. In the 2024 competition, most top-performing solutions were based on appearance-based  MCPT tracking frameworks~\cite{yoshida2024overlap,kim2024cluster}. These methods typically used clustering techniques—such as hierarchical clustering—to group tracklets based on visual similarity. Many also addressed challenges arising from visually ambiguous detections, such as occlusions or uncommon appearances using pose estimation.

In contrast, our approach aims to remain simple yet effective and adaptable to multi-class targets. Leveraging the fact that the camera views are overlapping, we first estimate a set of globally unique tracklets. Using camera calibration parameters, we project 2D bounding boxes into world coordinates and then perform progressive global ID assignment. Specifically, local tracklets from individual camera views are matched and assigned to global tracklets based on both trajectory similarity and appearance features, allowing for continuous updates and consistent global identity tracking.

\section{Approach}

Our approach is designed to be simple, generalizable, and easy to deploy, avoiding task-specific engineering such as pose estimation or object-specific clustering pipelines. Instead, we focus on building a robust yet lightweight tracking system that performs well across various object types and camera setups.

We begin by retraining a YOLOX-based object detector using the official training set provided in the 2025 AI City Challenge. To ensure generalization across all object categories of interest, we specifically select two scenes that contain annotated instances of all six target classes: \texttt{Person}, \texttt{Forklift}, \texttt{NovaCarter}, \texttt{Transporter}, \texttt{FourierGR1T2}, and \texttt{AgilityDigit}. This selection ensures that the retrained detector can perform consistently across diverse object appearances and motion patterns.

After detection, we extract feature embeddings for each bounding box using a standard ReID network. This extractor is class-agnostic and shared across all categories, enabling consistent appearance encoding for all tracked objects, regardless of class or view. By avoiding class-specific tuning, this design supports flexible deployment in real-world multi-object tracking scenarios.

\begin{figure}[t]
    \centering
    \includegraphics[width=\columnwidth]{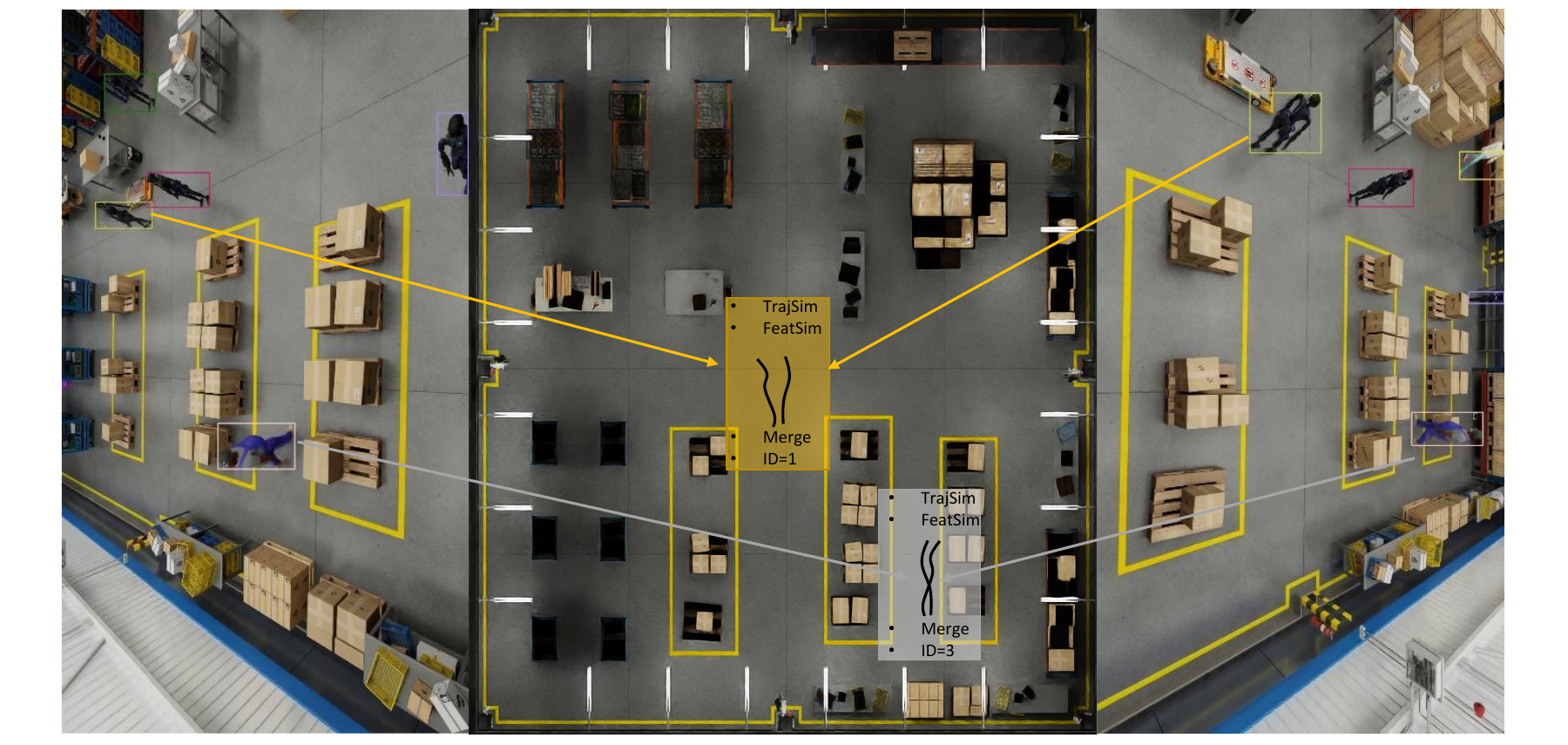}
    \caption{Glance initialization. Trajectory similarity is first computed in world coordinates, followed by feature similarity based on appearance. If both criteria are satisfied, the corresponding tracklets across views are assigned a shared global ID.}
    \label{fig:glance}
\end{figure}

\subsection{Single-Camera Tracking and Post-Processing}
For each individual camera view, we perform single-camera multi-object tracking using BoT-SORT~\cite{aharon2022bot}, a lightweight and effective tracker that combines Kalman filter-based motion prediction with appearance matching via ReID embeddings. By integrating both motion and visual cues, BoT-SORT produces robust tracklets per view.

To improve the consistency and compactness of these tracklets, we apply two post-processing steps.

\begin{figure*}[t]
    \centering
    \includegraphics[width=\textwidth]{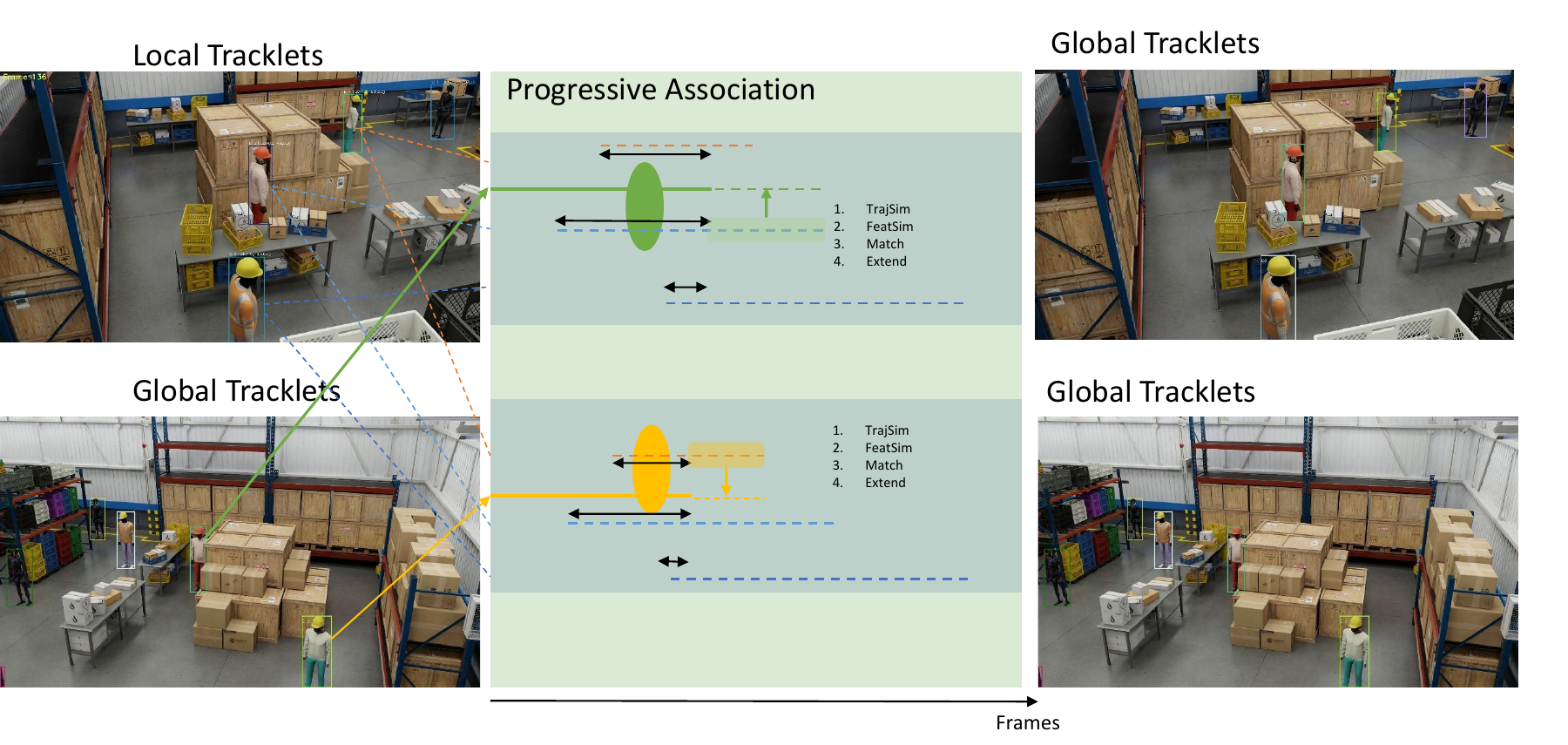}
    \caption{Illustration of the global association strategy across multiple overlapping views. Each local tracklet is matched to a global ID based on trajectory similarity in the shared frames and representative feature similarity. Then the match local tracklet is attached to the global tracklet and inherits a shared global ID.}
    \label{fig:global_association}
\end{figure*}

\textbf{Sequential Non-Max Suppression.}
The first stage involves merging redundant or fragmented tracklets within each camera view. We evaluate all tracklet pairs with temporally overlapping frames and compute their average spatial IoU. If this value exceeds a predefined threshold, the tracklets are merged to suppress duplication caused by occlusion, detection failure, or identity switches.

To formalize this, we adopt a Sequential NMS (SNMS) approach. Unlike standard NMS, which may underperform under scale variation or partial occlusion, SNMS uses a combination of temporal and spatial overlap. The spatial overlap is measured using the overlap coefficient:
\begin{equation}
\mathrm{OC}(A, B) = \frac{|A \cap B|}{\min(|A|, |B|)},
\end{equation}
where $A$ and $B$ are the sets of bounding boxes shared by two tracklets in overlapping frames. Tracklet pairs that exceed both spatial and temporal thresholds are either merged or reduced by removing the smaller one.

\textbf{Representative Features.}
While the previous step refines tracklet structure by eliminating spatial redundancies, appearance features across a tracklet can still vary significantly due to occlusions, partial views, or extreme viewpoints. To enhance the reliability of visual representations, we next select a compact set of representative features for each tracklet.

Our motivation stems from the observation that most frames in a tracklet capture typical or unobstructed views, while only a minority suffer from distortions. By identifying the dominant cluster of consistent features, we aim to isolate frames that best characterize the object's appearance. Specifically, we apply \texttt{DBSCAN()} clustering over the normalized ReID embeddings associated with each tracklet, using cosine distance as the similarity metric. The algorithm identifies dense regions in the embedding space, and we select the largest cluster as the dominant appearance group. From this cluster, a fixed number of samples (e.g., 20) are chosen to serve as the representative features.
This strategy helps suppress noisy or outlier features and improves the robustness of global association in subsequent cross-camera matching stages.

\subsection{Multi-Camera Tracking}

Our multi-camera tracking framework operates in two stages. In the initial \textit{glance initialization} phase, we examine a short time window across all camera views to estimate the set of unique global tracklets. This forms a reference set representing distinct objects seen from multiple viewpoints. In the second phase, we perform \textit{progressive association}, where local tracklets from each camera are incrementally matched to the global tracklets based on spatial and appearance similarity, continuously refining the global tracking result.

\subsubsection{Glance Initialization}

In the first stage of our multi-camera tracking framework, we perform \textit{glance initialization} by observing a short time window from all camera views. Due to the overlapping field-of-view among cameras, we assume that within this brief initial period, every object in the environment appears in at least one view. This assumption allows us to construct a complete set of initial global tracklets, representing the unique objects present in the scene.

To identify these unique global tracklets across views, we perform a two-step association:

\begin{itemize}
    \item \textbf{Trajectory Similarity:} We first use camera calibration matrices to project each initial tracklet's bounding boxes into world coordinates. Based on the resulting trajectories, we compute spatial proximity and motion similarity to evaluate whether two tracklets across different views could correspond to the same object.
    
    \item \textbf{Appearance Similarity:} In parallel, we compare the representative features of each tracklet—extracted during the single-camera post-processing stage—using cosine similarity. Only tracklets with high visual similarity are considered matching candidates.
\end{itemize}

If both trajectory and appearance similarities meet predefined thresholds, we merge the involved tracklets under a single global ID. Otherwise, a new global tracklet is initialized. The glance initialization process is shown in Figure \ref{fig:glance}. After the glance phase, we fix the number of global tracklets and only allow new global IDs under rare, conservative conditions to prevent identity fragmentation in subsequent steps.

% \begin{figure*}[t]
%     \centering
%     \includegraphics[width=\textwidth]{Figures/Glance_Fig.pdf}
%     \caption{Illustration of the global association strategy across multiple overlapping views. Each local tracklet is matched to a global ID based on trajectory and appearance similarity.}
%     \label{fig:global_association}
% \end{figure*}

\subsubsection{Progressive Association}

After initializing the global tracklets in the glance phase, we proceed to the second stage: \textit{progressive association}. In this phase, we aim to associate the remaining local tracklets from each camera view to the previously established global tracklets. We assume that every local tracklet corresponds to one of the global identities, as long as they share common temporal frames.

To perform the association, we first group local tracklets based on the overlap in time with existing global tracklets. For each candidate pair of local and global tracklets with shared frames, we compute:

\begin{itemize}
    \item \textbf{Trajectory Similarity:} Using world-coordinate projections (from camera calibration), we compare the trajectory segments that overlap temporally. This localized comparison ensures accurate geometric alignment.
    \item \textbf{Appearance Similarity:} We compute cosine similarity between the representative feature sets extracted for both the local and global tracklets.
\end{itemize}

Unlike one-to-one matching schemes such as the Hungarian algorithm, our approach allows local tracklets to be associated with multiple global tracklets if it is observed from different camera views. However, we impose a strict constraint that each global ID can only appear once per view at any given time, preventing the same identity from being duplicated in a single camera.
Once a local tracklet is associated with a global identity, we update the corresponding global tracklet by extending its trajectory and appending new representative features. This process is repeated progressively, associating newly available local tracklets with updated global tracks, until all local tracklets are assigned and no unassociated segments remain. A demo of the progressive association is shown in Figure \ref{fig:global_association}.

At the end of this process, a few local tracklets may still remain unassigned due to insufficient trajectory or appearance similarity. To handle such cases, we consider two alternative strategies:  
(i) Forced matching, where each remaining local tracklet is assigned to the most compatible global tracklet regardless of similarity thresholds, based on closest temporal and spatial proximity;  
(ii) Global ID expansion, where we treat one of the unassigned local tracklets—specifically, the longest—as a new global tracklet. This increases the chance of frame overlap in subsequent iterations, after which we resume the progressive association process with the updated global tracklet pool.

\section{Experiments}

\subsection{Dataset Overview}

We evaluate our method on the MTMC\_Tracking\_2025 dataset from NVIDIA's Physical AI Smart Spaces initiative. This synthetic dataset, generated using NVIDIA Omniverse and Isaac Sim, offers a comprehensive benchmark for MTMC tracking in complex indoor environments such as warehouses, hospitals, and laboratories. The dataset encompasses over 250 hours of video captured from nearly 1,500 cameras, providing time-synchronized, high-resolution (1080p at 30 FPS) footage across diverse scenes.

The MTMC\_Tracking\_2025 subset includes 23 training and validation scenes and 4 test scenes, each accompanied by detailed ground truth annotations in JSON format. These annotations provide 2D and 3D bounding boxes, object identities, and global coordinates, facilitating robust evaluation of tracking algorithms. The dataset features a variety of object classes, including \texttt{Person}, \texttt{Forklift}, \texttt{NovaCarter}, \texttt{Transporter}, \texttt{FourierGR1T2}, and \texttt{AgilityDigit}, totaling 363 unique objects in the train/val split. Calibration files for each camera view are provided to enable accurate projection of detections into world coordinates.

\subsection{Evaluation Metric}

The MCMT Track of the AI City Challenge evaluates submissions based on the Higher Order Tracking Accuracy (HOTA) metric~\cite{luiten2021hota}. HOTA addresses the limitations of earlier metrics such as MOTA and IDF1 by offering a unified evaluation framework that jointly considers detection accuracy, identity association quality, and localization precision. Specifically, HOTA is computed as the geometric mean of Detection Accuracy (DetA) and Association Accuracy (AssA), providing a balanced view of both detection and identity consistency across frames.

In this year's challenge, evaluation is performed in 3D world coordinates. Each detection is represented by its centroid $(x, y, z)$, dimensions, and yaw angle in meters and radians, respectively, rather than 2D image-space bounding boxes. This highlights the need for accurate camera calibration and 3D localization across views.

\subsection{Implementation Details}

We trained the YOLOX-X~\cite{ge2021yolox} detector on two training scenes from the AI City Challenge 2025 dataset. The detector was initialized with weights pre-trained on MOT17, CrowdHuman, CityPersons, and ETHZ datasets, following the setup used in ByteTrack~\cite{zhang2022bytetrack}. For ReID feature extraction, we employed OSNet-AIN~\cite{zhou2019omni}, trained on Market1501, CUHK03, and MSMT17 datasets to ensure generalization across diverse appearances.

To further improve the HOTA score, we applied a tracklet interpolation step (similar to \cite{hashempoor2025deep}) to fill in short gaps of missing detections. Rather than interpolating all missing segments—which risks introducing bounding boxes outside the camera's field of view—we only interpolate when the number of consecutive missing frames remains below a fixed threshold. This selective strategy helps complete fragmented tracklets while minimizing false positives from over-extension. Implementation is avalable at:
\href{https://github.com/Hamidreza-Hashempoor/Glance-MCMT}{\textcolor{magenta}{https://github.com/Hamidreza-Hashempoor/Glance-MCMT}}

\subsection{Results}

We evaluated our proposed algorithm on the validation set of the AI City Challenge 2025 and taking average in Table \ref{tab:results}, using two distinct strategies during the progressive association stage: (i) Forced Matching (FM), where unassigned local tracklets are forcibly matched to the closest global tracklets regardless of similarity scores; and (ii) Global ID Expansion (GIDE), where unmatched local tracklets are used to initialize new global identities when no suitable association is found. The comparison of these strategies allows us to analyze the trade-offs between conservative matching and identity expansion in maintaining global track consistency.

\begin{table}[ht]
\centering
\begin{tabular}{lcccc}
\toprule
\textbf{Method} & \textbf{HOTA} & \textbf{DetA} & \textbf{AssA} & \textbf{LocA} \\
\midrule
FM & 43.09 & 45.12 & 41.16 & 61.58 \\
GIDE & 51.34 & 48.19& 54.71 & 66.19 \\
\bottomrule
\end{tabular}
\caption{Performance comparison of our method FM and GIDE progressive association strategies.}
\label{tab:results}
\end{table}

\section{Discussion}

\subsection{Accuracy and HOTA Values}

The current version of our solution, utilizing lightweight modules and basic models, provides competitive performance in terms of HOTA. However, there is significant potential for improvement. One promising direction is the incorporation of more advanced feature engineering techniques. Similar to recent Track1 papers in the AI City Challenge, integrating more advanced feature clustering and pose estimation during re-identification could enhance identity association and improve HOTA scores, especially in challenging scenarios like occlusions or non-frontal views.

Another aspect that can be optimized is the pixel-to-world coordinate translation. Inaccurate 2D bounding boxes, often due to partial occlusions or unusual object orientations, lead to errors when mapping to 3D world coordinates. These inaccuracies result in trajectory mismatches. By employing a robust filtering techniques, one can reduce errors in world coordinate translation, resulting in more accurate trajectories and improved overall performance.

\subsection{Extension to Online Tracking}

Our approach can be easily extended to an online tracking scenario. In the online setting, we maintain a running history of tracklets to extract representative features as new tracklets are generated. At the beginning, we still perform the glance initialization step using a few frames from all camera views to establish the initial global tracklets. 
As new local tracklets appear in each camera view, they are progressively associated with the global tracklets based on trajectory and appearance similarity, similar to the offline process. If a new object enters the scene, i.e., a new local tracklet is generated, we perform a global matching step. During this step, we match the newly introduced object to a global ID that has already been assigned in the other views, ensuring consistency across the system.

\section{Conclusion}

In this paper, we presented a lightweight and efficient approach for  MCMT tracking. By combining simple yet effective methods such as glance initialization, progressive association, and representative feature extraction, we achieved competitive results with promising HOTA scores. Our method can be easily extended to online tracking, making it suitable for real-time applications. Future work will focus on incorporating advanced feature engineering and improving coordinate translation accuracy to further enhance performance.
{
    \small
    \bibliographystyle{ieeenat_fullname}
    \bibliography{main}
}

\end{document}